\documentclass[conference]{IEEEtran}
\IEEEoverridecommandlockouts
\usepackage{cite}
\usepackage{amsmath,amssymb,amsfonts,hyperref}
\usepackage{graphicx}
\usepackage{textcomp}
\usepackage{xcolor}
\usepackage{comment}
\usepackage{amsmath,amssymb} % define this before the line numbering.
\usepackage{color}
\usepackage[noend]{algpseudocode}
\usepackage{algorithmicx,algorithm}
\usepackage{multirow}
\usepackage{subfigure}

\usepackage{bm}

\algblock{Input}{EndInput}
\algnotext{EndInput}
\algblock{Output}{EndOutput}
\algnotext{EndOutput}

\def\BibTeX{{\rm B\kern-.05em{\sc i\kern-.025em b}\kern-.08em
    T\kern-.1667em\lower.7ex\hbox{E}\kern-.125emX}}

\begin{document}

\title{Tree-based Text-Vision BERT for Video Search in Baidu Video Advertising
}

\author{
\IEEEauthorblockN{Tan Yu,  Jie Liu, Yi Yang, Yi Li, Hongliang Fei, Ping Li} \\
\IEEEauthorblockA{Cognitive Computing Lab, Baidu Research} 
\IEEEauthorblockA{Baidu Search Ads (Phoenix Nest), Baidu Inc.} 
10900 NE 8th St. Bellevue, Washington 98004, USA\\
No. 10 Xibeiwang East Road, Beijing 100193, China \\\\
\{tanyu01, liujie34,  yangyi15, liyi01, hongliangfei, liping11\}@baidu.com
}

\maketitle

\begin{abstract}
\footnote{This revision is based on a manuscript submitted in October 2020, to  ICDE'21. We thank the Program Committee for their valuable comments.}The advancement of the communication technology and the popularity of the smart phones foster the booming of video ads.  Baidu, as one of the leading search engine companies in the world,   receives billions of search queries per day. How to pair the video ads with the user search is the core task of Baidu video advertising.  Due to the modality gap, the query-to-video retrieval is much more challenging than traditional query-to-document retrieval and image-to-image search. Traditionally, the query-to-video retrieval is tackled by the query-to-title retrieval, which is not reliable when the quality of tiles are not high. With the rapid progress achieved in computer vision and natural language processing  in recent years,  content-based search methods  becomes promising for the query-to-video retrieval.  Benefited from pretraining on large-scale datasets, some visionBERT methods based on cross-modal attention have achieved excellent performance in many vision-language  tasks not only in academia but also in industry.  Nevertheless, the expensive computation cost of cross-modal attention makes it impractical for large-scale search in industrial applications. In this work, we present a tree-based combo-attention network (TCAN) which has been recently launched in Baidu's dynamic video advertising platform.  It provides a  practical solution to deploy the heavy cross-modal attention  for the large-scale query-to-video search.  After launching tree-based combo-attention network, click-through rate gets improved by $2.29\%$ and conversion rate get improved  by $2.63\%$.\\
\end{abstract}

\begin{IEEEkeywords}
advertising, search, cross-modal
\end{IEEEkeywords}

\section{Introduction}

  Since high-quality videos can quickly build the engagement with the audience, video ads are substantially more  compelling over its counterparts. 
Recently,  with the popularity of  reliable high-speed internet,  videos can be transferred in seconds, which fosters the blooming of video advertising market. %The advertisers are pouring more 
The advertisers are putting more efforts on video advertising to build a relationship with customers in a more effective way. Baidu, as one of leading search engine companies of the world, receives billions of text queries from users' searches. Linking the relevant video ads provided by advertisers with  potential customers according to their search queries is the main task  of Baidu video advertising.  
The quality of matching the query with the video ads directly influences the revenue of the advertisers.
%The linking quality will significantly influence 

In essence, linking the video as with the user query is a cross-modal retrieval problem, as the query is in the text modal, and the ads are in the video modal.   Due to modal gap, the cross-modal retrieval is much more challenging than traditional query-to-document retrieval in existing mainstream search engine.  Traditionally,  the query-to-video retrieval is converted into text-to-text retrieval by matching the query text with the video title. Since both video titles and user queries are text,  it can be readily addressed by current text retrieval model. Nevertheless, it requires expensive human labors to create the high-quality titles for videos. Meanwhile, the manually-labeled titles are subjective to the annotators,  and thus might not effectively embody the visual content.    To improve the quality of the text-to-visual retrieval, a more reliable way is to directly match the user's query with the visual content through the natural language processing (NLP) and computer vision techniques.

 In the past years, we have witnessed rapid progress in both computer vision and NLP. Some deep learning models pretrained by large-scale dataset have achieved significantly better performance than traditional methods based on hand-crafted features. To be specific,  convolutional neural network (CNN)~\cite{he2016deep} has  substantially improved the performance in image/video recognition.  Meanwhile, its output of a CNN's hidden layer is an effective image/video representation which can be used for visual-to-visual retrieval.  In parallel, the transformer~\cite{vaswani2017attention} and BERT~\cite{devlin2019bert} has  achieved substantial success in many NLP tasks.      Similarly, the output of a BERT's hidden layer is also effective for text-to-text retrieval.    Despite the CNN feature and BERT feature have achieved significant success in visual-to-visual and text-to-text retrieval tasks, the text-to-visual search in our task is still challenging.

 \begin{figure*}[t]
    \centering
    \includegraphics[width=6.5in]{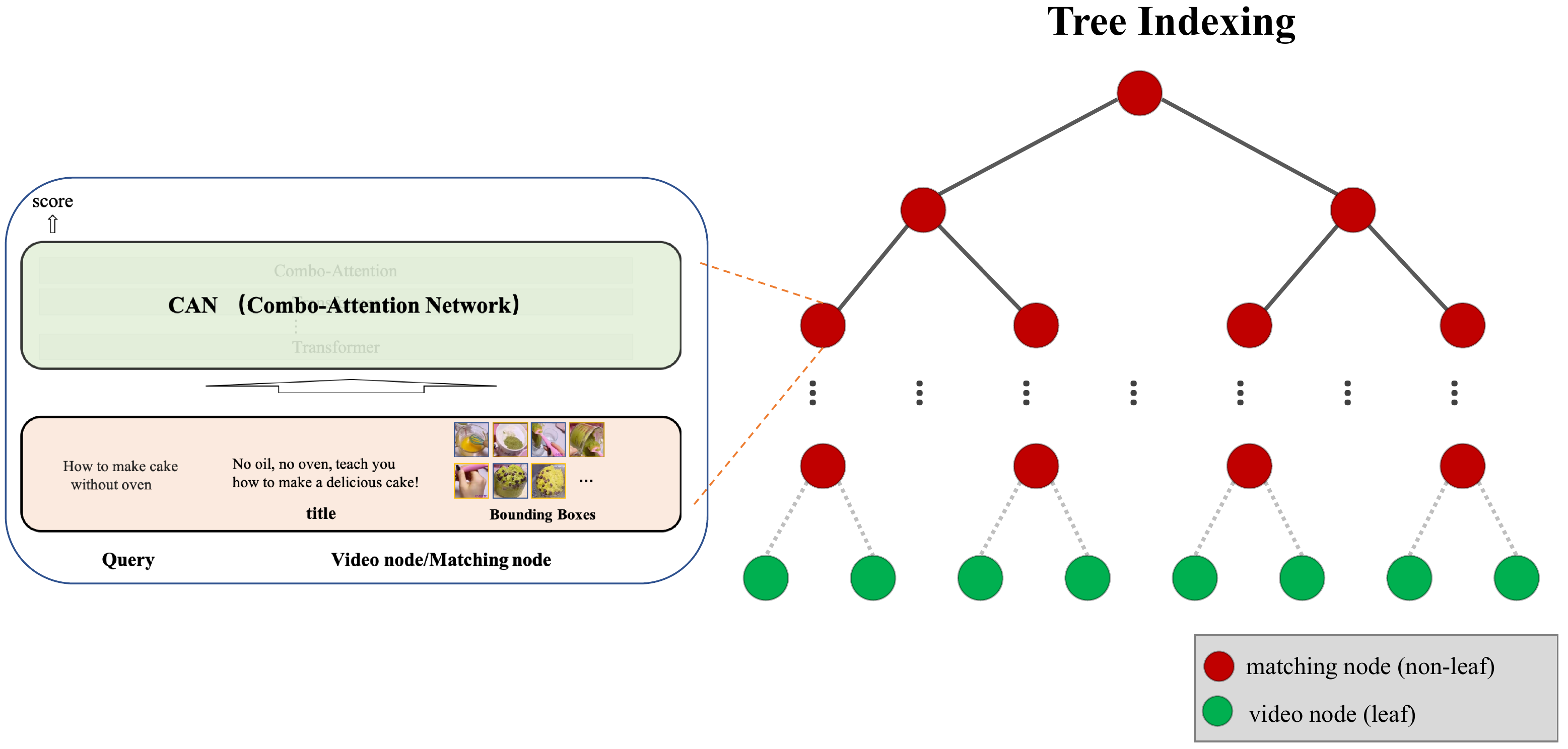}

    \caption{The architecture of the proposed tree-based combo-attention network (TCAN). In the retrieval phase, it only uses CAN to calculate the  similarity between the query and the tree node along the traverse path.  Since the number of traversed nodes is significantly smaller than that of all videos in the database, the efficiency is considerably boosted. }
    \label{architecture}\vspace{-0.1in}    
\end{figure*}

 Traditionally, the text-to-visual cross-modal retrieval is solved by the joint embedding~\cite{farhadi2010every,frome2013devise}.  It maps the features from different modals into the same feature space and thus their similarity can be directly measured by their Euclidean distance in the joint feature space.  In the training phase, it seeks to enlarge the distance between the text and its irrelevant images/videos in the joint feature space, and meanwhile minimize the distance between the text and its relevant images/videos in the joint feature space.  One of the amazing properties of the joint embedding is that it generates the global text/visual features, and is well compatible with  approximate nearest neighbor  (ANN) search techniques such as inverted indexing or graph/tree-based indexing.  It is hence quite efficient for large-scale cross-modal retrieval applications.

 %On the other hand, global features might 

 %Neverthless, 
 
 %When optimizing the 

Nevertheless, the global features used in joint embedding methods might not be able to conduct fine-level matching between words and local regions of an image/video.  In many cases, only few words in the query sentence are relevant with some small local regions in the video/video. Therefore, to conduct local matching more effectively, some methods~\cite{lee2018stacked}  rely on local features. Basically, they represent an image by a set of bounding box features and represent the query sentence as a set of word features. Then the relevance score is determined by the set-to-set matching.  
 Recently, inspired by the breakthrough achieved by BERT  in NLP, many vision-based BERT methods~\cite{sun2019videobert, lu2019vilbert,yu2021multi,liu2021inflate} have been proposed, and achieve an excellent performance in cross-modal tasks like visual question answering (VQA), image captioning and cross-modal retrieval.  In parallel, Baidu has launched a combo-attention network (CAN) ~\cite{yu2020combo} for an effective query-to-video retrieval in dynamic video adverting (DVA)  platform.  Nevertheless, since the cross-attention mechanism used in CAN takes expensive computation cost,  it is impossible to use CAN to compute the similarity between the query and all video ads due to limited computing resource. Thus, previously, we first conducted the coarse-level retrieval through title-based retrieval and deployed CAN in the re-ranking phase for the search efficiency.   Nevertheless, in this case, some relevant videos might be filtered out in the title-based retrieval due to their low-quality titles.  A more reasonable way is to incorporate the cross-modal attention in the early stage.
 
 %the visual information is only considered until the re-ranking phase, 

 Recently, tree model is revisited  to speed up the deployment of deep models in the recommendation system.  Nevertheless, the query's  feature used in our CAN is a set of discrete local features, and thus cannot be represented as an embedding feature  to build a tree as TDM~\cite{zhu2018learning,yu2022egm}.  To tackle this problem, we propose a novel dual-path CAN to make it compatible with tree-based model. Based on it,  we build Tree-based Combo-Attention Network (TCAN), which is  recently launched in Baidu's dynamic video advertising platform.  It  achieves high effectiveness by using the combo-attention network (CAN), and meanwhile achieves high efficiency by exploiting the tree structure which avoids exhaustively comparing with each sample in the database.  
We visualize the overview of the proposed TCAN in Figure~\ref{architecture}.  Basically, given a query, it traverses the  tree from the root to leafs, the combo-attention network is only computed on nodes along the traverse trace, which avoid exhaustively comparing  the query with all videos  in the database. Given $N$ reference videos in the database, the tree model decreases the computation complexity  from $\mathcal{O}(N)$ to $\mathcal{O}(\mathrm{log}(N))$, making it feasible for online serving. After launching the TCAN, we achieve a $2.26\%$ increase in CTR,  and a $2.63\%$ increase in CVR.

In a nutshell, the contributions of this paper are four-fold:
\begin{itemize}
   \item We upgrade the two-stage deployment of original CAN to a one-stage CAN structure. The one-stage CAN simultaneously  considers the video's title and its visual content.    
    \item We design a dual-path structure for one-stage CAN,  which supports feature embedding as well as cross attention simultaneously.
    \item  Based on the proposed dual-path CAN, we built a tree-based CAN. Benefited from the tree structure, it achieves an effective and efficiency cross-modal retrieval.  
    \item We have validated the effectiveness of tree-based CAN in Baidu's dynamic video advertising platform.   The promising results demonstrate its usefulness  for large-scale cross-modal retrieval in industrial applications.
\end{itemize}

\section{Related Work}
We review the related work in four fields: cross-modal retrieval, approximate nearest neighbor search, fast neural network and Baidu's search ads.

\subsection{Cross-modal Retrieval}

The traditional cross-modal retrieval tasks are tackled by joint embedding  ~\cite{farhadi2010every,frome2013devise}. It maps the images and texts from their original feature space into a joint feature space so that they can be compared directly.  They target to learn a mapping function, which maximizes the distance  between an irrelevant image-text pair to be large , and meanwhile minimizes the distance between relevant image-text pairs. To learn the mapping function more effectively.  VSE++~\cite{faghri2018vse++} conducts hard negative sampling, which only penalizes the hardest negative sample in each mini-batch.  Since the joint-embedding methods generate global features, it well supports feature indexing and thus is quite suitable for large-scale cross-modal retrieval. On the other hand, the global features cannot model the fine-level matching between words and local regions, and thus it might not be effective to capture relevance between an image and a text sentence. 

To more effectively describe the relevance between the image and text, several works~\cite{lee2018stacked} rely on local features.  To be specific, they represent an image by a set of bounding box features which are extracted by pre-trained object detectors. The detected bounding boxes are the candidate locations of objects in the image.  Meanwhile, they represent the sentence by a set of word features.  Then the relevance between the image and the text sentence is obtained by set-to-set matching between word features and bounding box features.  When conducting the set-to-set matching, the relevant word-box pairs with a large similarity score can be easily identified.

To further improvement the performance of  text-vision retrieval methods. Several work ~\cite{de2017modulating,ma2015multimodal,yu2017end} inject the cross-modal context when generating image/text representation. For instance,  when computing the sentence representation,   m-CNN~\cite{ma2015multimodal} concatenate the image's feature with words' features  as the input of the 1D convolution. 
In parallel,   when computing image features, BCN~\cite{de2017modulating} uses text features  to guide the generation of image features.   Recently, inspired by the triumph achieved by BERT in many natural language processing (NLP) tasks~\cite{devlin2019bert}, some vision-BERT methods ~\cite{sun2019videobert,lu2019vilbert}   are proposed, achieving excellent performance in many cross-modal tasks such as visual question answering (VQA), image/video captioning and cross-modal retrieval.  Basically, the vision-BERT method can be coarsely grouped into two categories. The methods of the first category adopt a single-stream structure. It treats the bounding box features  and word features without bias, and directly concatenate them as the input of the self-attention modules.   The methods of the second category adopt a two-stream structure.  In the text stream, the words features  generate query vectors, and the bounding box features are used to generate value vectors and key vectors. 
On the other hand, in the vision stream, the bounding box features  generate query vectors, and the word features are used to generate value vectors and key vectors.  In fact, both single-stream and two-stream vision-BERT methods are computationally costly. Therefore, it is impractical to directly use them to conduct large-scale cross-modal retrieval.

\subsection{Approximated Nearest Neighbor Search}
To boost the search efficiency, many approximated nearest neighbor (ANN) search methods are proposed.   The research on ANN dates back to at least the 1970s~\cite{friedman1975algorithm,friedman1977algorithm}.   Traditionally, ANN search methods mainly include hashing-based methods~\cite{indyk1998approximate,datar2004locality,li2005using, weiss2008spectral,shrivastava2012fast,li2014coding,wei2014scalable,li2019sign}, quantization-based method~\cite{jegou2011product,ge2013optimized,zhang2014composite,andre2015cache},  tree-based method~\cite{lin1994tv,berchtold1996x,zezula1998approximate,sakurai2000tree},  and graph-based method~\cite{fu2019fast,tan2019efficient,malkov2020efficient,zhao2020song}. In recent years,  ``neural ranking'' has also attracted increasingly more attentions~\cite{zhu2018learning,zhu2019joint,tan2020fast,zhuo2020learning,gao2020deep,tan2021fast,yu2022egm}.  Also, inspired by the success of deep learning, many deep Hashing~\cite{erin2015deep,zhu2016deep} and deep PQ~\cite{cao2016deep,yu2018product} work are proposed. Basically, them incorporate the Hashing and PQ in a neural network, and trains the Hashing or PQ codes in an end-to-end manner.

Despite Hashing codes and PQ codes can enable the fast computation between the query and each reference item, when the number of reference items in the database is large, the time cost is still costly. 
To make the retrieval more scalable to large-scale dataset, some non-exhaustive search methods are proposed.  Inverted multiple indexing  (IMI)~\cite{babenko2015inverted} is one of the most popular methods for non-exhaustive search.   Basically, it partitions the feature space into fine cells through the k-means.  It only computes the similarity between the query and the cell centroids.   Since the number of centroids is significantly smaller than the number of reference items, the efficiency is considerably boosted. In parallel, K-Dimensional tree~\cite{sproull1991refinements} is another widely used strategy for avoiding comparing the query with each item exhaustively. It builds a binary search tree to partition the feature space, and formulates the retrieval into a tree traverse problem.  Recently, TDM~\cite{zhu2018learning} also exploits the tree structure to boost the efficiency of deep models deployed in recommendation system. The core idea is that, it only evaluate the expensive deep models  in a small subset of tree nodes and avoids evaluating the deep models on all items in the database.

\subsection{Fast Neural Network}

The heavy computation cost of deep neural network limits its usefulness in online serving or mobile applications.  Many studies~\cite{han2015deep} have shown that, network weights might be redundant and do not convey much information. In some cases~\cite{zhang2017understanding}, some large-scale models tend to memorize the dataset instead of learning some generic capability, suffering from serious over-fitting.  To boost the efficiency and suppress over-fitting, recently, many efficient and compact neural network architecture are developed, achieving very promising performance in both effectiveness and efficiency. One of most widely used strategy is training  quantized neural networks~\cite{courbariaux2015binaryconnect, rastegari2016xnor, ott2016recurrent, wu2016binarized, zhang2017zipml,hubara2017quantized}. For instance,~\cite{courbariaux2015binaryconnect,rastegari2016xnor}  design the binary neural network where the value of weights are chosen from two constants. In parallel, some efficient neural networks~\cite{han2015deep,wen2016learning, iandola2016squeezenet} are obtained by compressing already-trained neural networks.  Inspired by the recent success achieved by knowledge distillation~\cite{hinton2015distilling}, some methods~\cite{polino2018model,jiao2020tinybert} propose to find a more compact student network through knowledge distillation. To be specific,~\cite{polino2018model} jointly learn the weight quantization and the model distillation.~\cite{jiao2020tinybert}  designs a Tiny-BERT by distilling the knowledge from the large-scale BERT model. Following the spirit of Tiny-BERT, in our work, we also exploit the knowledge distillation for model compression.
%constrain the network weights to be binary code.

%Since deep neural network are normally over-parameterized, many methods seek to compress the model to achieve a higher efficiency. 

\subsection{Baidu's Search Ads}

Baidu Search Ads (a.k.a. ``Phoenix Nest'') is one of the most important revenue sources for Baidu~\cite{fan2019mobius,fei2021gemnn,xu2021agile}. In the search industry, sponsored online advertising produces many billions of dollar revenues for online advertisers.  Since around 2017, Baidu Search Ads has been undergoing several major upgrades by incorporating the rapid-growing technologies in near neighbor search, machine learning, and  systems. For example,~\cite{zhao2019aibox} reported new architectures for distributed GPU-based parameter servers which have replaced the MPI-based system for training CTR models.~\cite{fan2019mobius} described the widespread use of approximate near neighbor search (ANNS) and maximum inner product search (MIPS)~\cite{zhou2019mobius,zhao2020song} to  substantially improve the quality of ads recalls in the early stage of the pipeline of  Baidu's ads system.   In recent years, Baidu's short-form video recommendations~\cite{li2020video} and video-based search ads have achieved great progress~\cite{yu2020combo,yu2021multi,yu2022egm}. In this paper, we introduce  the technology for a representative  project which has significantly boosted Baidu's video-based ads revenues.

%Considering the efficiency, in many cases, we cannot conduct the 
%Approximated nearest neighbor search is 

%Besides the effectiveness, the efficiency is the other key point for a retrieval system. 
% ViLBERT~\cite{lu2019vilbert} further improves VideoBERT by designed a two-stream architecture consists of a text stream and an image stream. In each stream, they design a co-attention transformer layer which takes both two modals as input to generate the attention.
%Similarly, MCN~\cite{yu2019deep} also uses the text feature to guide the attention when generating the feature of the image.  Nevertheless,  both ViLBERT and MCN are designed for the image-text tasks, which can not be directly used for video-text tasks. 
%  Visual captioning~\cite{} is a dual task with respect to the text-to-vision retrieval task. Different from text-to-vision retrieval which uses text query to retrieve the visual items, Visual captioning takes the an image/video as input to generate a sentence describing the content in an image or video.   To be continued ....
%\subsection{Visual Captioning}
%
%\begin{figure}[t!]
%\centering
%    \subfigure[Self-Attention]{\includegraphics[width=0.20\textwidth]{figures/sanew.pdf}}  
%    \hspace{2mm}
%    \subfigure[Combo-Attention]{\includegraphics[width=0.20\textwidth]{figures/caself.pdf}}  
%    \caption{The architecture of the self-attention (SA) module and the combo-attention (CA) module.}
%    \label{basicmodule}
%\end{figure}

\section{Background}
\label{cansec}
%Before we introduce the  tree-based Combo-Attention Network, we first briefly introduce the combo-attention network (CAN).

%In this section, we introduce the proposed  combo-attention network (CAN).

\subsection{Definition}

Given a video, we extract  key frames from it. For each key frame,  bounding boxes are detected by Faster R-CNN.   Each detected bounding box denotes the location of a candidate object in the key frame.   Note that, the detected bounding boxes normally have overlap with each other, and thus there is significant redundancy among the detected bounding boxes. 
Therefore, after we obtain all the bounding boxes of all keyframes, we utilize the k-means clustering to group them into $K$ clusters.  
 Then the video's representation is the  set of cluster centroids, $\mathbf{C} =[ \mathbf{c}_1, \cdots, \mathbf{c}_K]$.  Given a query sentence, we obtain  a sequence of word features  $\mathbf{W} = [\mathbf{w}_1,\cdots,\mathbf{w}_{M}]$ through a pretrained word embedding. 

\subsection{Basic Block}

The combo attention network takes $\mathcal{W}$ and $\mathcal{C}$ as input. Basically, it concatenates $\mathcal{W}$ and $\mathcal{C}$ into a new sequence:
\begin{equation}
\mathbf{M}_{0} = [\mathbf{c}_1, \cdots, \mathbf{c}_K, \mathbf{w}_1,\cdots,\mathbf{w}_{M}].
\end{equation}
It adopts  a series of standard self-attention modules to process the combined sequence. To be specific, let denote  the input  of the $i$-th self-attention module by $\mathbf{M}_{i-1}$  and denote the 
output  of the $i$-th self-attention module by $\mathbf{M}_{i}$. It first computes the query vectors $\mathbf{Q}_i  = [\mathbf{q}_{i,1},\cdots,\mathbf{q}_{i,M+K}]$, key vectors $\mathbf{K}_i  = [\mathbf{k}_{i,1},\cdots,\mathbf{k}_{i,M+K}]$ and value vectors $\mathbf{V}_i  = [\mathbf{v}_{i,1},\cdots,\mathbf{v}_{i,M+K}]$ by 
\begin{equation}
    \mathbf{V}_i = f_i(\mathbf{M}_i),~\mathbf{K}_i = g_i(\mathbf{M}_i),~ \mathbf{Q}_i = h_i(\mathbf{M}_i),    
\end{equation}
where 
\begin{equation}
     f_i(\mathbf{M}_i)= \mathbf{W}_{f_i}\mathbf{\mathbf{M}_i} ,~~g_i(\mathbf{M}_i) =  \mathbf{W}_{g_i}\mathbf{\mathbf{M}_i},~~h_i(\mathbf{M}_i) =  \mathbf{W}_{h_i} \mathbf{\mathbf{M}_i},
\end{equation}
, and $\mathbf{W}_{f_i}$, $\mathbf{W}_{g_i}$ and $\mathbf{W}_{h_i}$ are weight matrices to be learned.  In implementation,  $ f_i(\cdot)$,   $g_i(\cdot)$ and $h_i(\cdot)$ are implemented by fully-connected layers with the bias fixed as $0$.

 For each query vector,  $\mathbf{q}_{i,j}$,  the self-attention module computes  the matrix-vector product between $\mathbf{q}_{i,j}$ and the key matrix $\mathbf{K}_i$ followed by a softmax operation,  the soft-attention vector $\mathbf{a}_{i,j}$ is obtained by 
\begin{equation}
    \mathbf{a}_{i,j} =\mathrm{softmax}(\mathbf{K}_i\mathbf{q}_{i,j}^{\top}),
\end{equation}
where $\mathrm{softmax}$ is defined as 
\begin{equation}
\mathrm{softmax}([x_1,\cdots,x_D]) =[\frac{e^{x_1}}{\sum_{i=1}^D e^{x_i}},\cdots,\frac{e^{x_D}}{\sum_{i=1}^D e^{x_i}} ].
\end{equation}

Then the attended feature vector $\mathbf{f}_{i,j}$ is obtained by a weighted summation over all columns of $\mathbf{V}_i$ and the weights are items in $\mathbf{a}_{i,j}$:
\begin{equation}
    \mathbf{f}_{i,j} =  \mathbf{V}_i \mathbf{a}_{i,j}^{\top}.
\end{equation}
The attended feature matrix $\mathbf{F}_i$ consists of all attended features:
\begin{equation}
    \mathbf{F}_i = [\mathbf{f}_{i,1},\cdots,\mathbf{f}_{i,K+M}].
\end{equation}
$\mathbf{F}_i$ goes through an add$\&$norm layer and generates:
\begin{equation}
   {\mathbf{M}}_i = \mathrm{norm}(\mathbf{F}_i + \mathbf{M}_{i-1}), 
\end{equation}
where $\mathrm{norm}(\cdot)$ denotes the layer normalization~\cite{ba2016layer}.  Note that, for easiness of illustration, the above formulation is based on the single head. In implementation, we adopt an $8$-head settings for all attention blocks.

\subsection{Similarity and Loss Function}

\textbf{Similarity computation.}  After the processing of $N$ layers of self-attention modules, we generate the self-attended features $\mathbf{M}_N = [\mathbf{m}_{N,1},\cdots,\mathbf{m}_{N,K+M}]$. Among them, $[\mathbf{m}_{N,1},\cdots \mathbf{m}_{N,1}]$ correspond to  the attended bounding box centroids and $[\mathbf{m}_{N,K+1},\cdots \mathbf{m}_{N,K+M}]$ correspond to the attended word features. We denote by $\mathbf{c}_{N,j} = \mathbf{m}_{N,j}$ as the $j$-th self-attended centroid and denote by  $\mathbf{w}_{N,j} = \mathbf{m}_{N,K+j}$ as the $j$-th self-attended word feature.

Soft attention layer $\mathbf{C}_N = [\mathbf{c}_{N,i},\cdots, \mathbf{c}_{N,K} ]$ as well as $ \mathbf{W}_N = [\mathbf{w}_{N,1},\cdots, \mathbf{w}_{N,M}]$ as input, and computes a similarity matrix $\mathbf{S}$ by
\begin{equation}
\label{beginW}
\mathbf{S} = {\mathbf{C}_N^{\top}}\mathbf{W}_N.
\end{equation}
%where $\mathbf{W} \in \mathbb{R}^{D_w\times N}$ corresponds to the matrix  containing all words features  $[\bar{\mathbf{w}}_1^{(2)},\cdots,\bar{\mathbf{w}}_M^{(2)}]$ of the sentence and 
%$\mathbf{B} \in \mathbb{R}^{D_b\times N}$ corresponds to the matrix  containing all bounding boxes features  $[\bar{\mathbf{b}}_1^{(2)},\cdots,\bar{\mathbf{b}}_N^{(2)}]$ of the image. 
 For each column of $\mathbf{S}$, $\mathbf{s}_i$, we conduct a soft-max operation on it and obtained a new vector:
\begin{equation}
\tilde{\mathbf{s}}_{i} = \mathrm{softmax}({\mathbf{s}}_{i} ).
\end{equation} 
Then  a new similarity matrix is obtained through $\tilde{\mathbf{S}} = [\hat{\mathbf{s}}_{1},\cdots,\hat{\mathbf{s}}_{M}]$. The output of  soft attention layer is computed~by  
\begin{equation}
\label{finalW}
\tilde{\mathbf{W}}_N =   \mathbf{C}_N \tilde{\mathbf{S}}.
\end{equation}
 The final similarity score is computed by  
 \begin{equation}
 \mathrm{s} = \sum_{i=1}^{M} \langle \tilde{\mathbf{w}}_{N,i},\mathbf{w}_{N,i} \rangle,
 \end{equation}
 where $\tilde{\mathbf{w}}_{N,i}$ denotes the $i$-th column of $\tilde{\mathbf{W}}_N$. 
 In the search phase, the obtained similarity score $s$ is used for ranking the videos given a text query. In the training phase, $s$ is used for constructing the training loss.

 \textbf{Training loss.} 
 The training is conducted on each mini-batch.  Each mini-batch consists of $K$ ground-truth video-sentence pairs $\{(V_k,S_k)\}_{k=1}^K$.  Each video  $V_k$ in the mini-batch is only relevant with the sentence in its ground-truth  sentence-video pair, $S_k$, and  irrelevant with other sentences.   In the training process,   we seek to maximize the similarities between relevant sentence-video pairs and minimize the similarities between irrelevant sentence-video pairs.   We denote   the similarity score of the video $V_i$ and the sentence $S_j$ by  $s(i,j)$, and construct the loss $\mathcal{L}$ by
 \begin{equation}
 \label{loss}
 \begin{split}
 \mathcal{L} &=  \sum_{k=1}^K  \sum_{j\neq k}[m - s(k,k) + s(k,j)]_{+} \\
 &+ \sum_{k=1}^K \sum_{j\neq k}[m - s(k,k) + s(j,k)]_{+} ,
 \end{split}
 \end{equation}
 where $[a]_{+}  = max(a, 0)$ is a clip function, and  $m$ is the margin which we set as $0.2$ by default. Based on the definition, the loss $\mathcal{L}$ only penalizes the pairs beyond the margin $m$. To some extent, this kind of settings only penalizes the hard negative samples.   Intuitively, the first part of $\mathcal{L}$ in Eq.~(\ref{loss}) focuses on the query side, which penalizes the hard negative videos with respect to the query. On the other hand, the second part of $\mathcal{L}$ focuses on the video side, which penalizes the hard negative queries with respect to the video.

 %Note that, the item $\sum_{j\neq k}[\alpha - s(k,k) + s(k,j)]_{+}$ in Eq.~(\ref{loss}) targets to make the similarity between $V_k$ and $S_k$ larger  by a margin $\alpha$ than  similarities between $V_k$  and other sentences in the mini-batch. In contrast, the item $\sum_{j\neq k}[\alpha - s(k,k) + s(j,k)]_{+} $ seeks to make the similarity between $V_k$ and $S_k$ larger  by a margin $\alpha$  than similarities between $S_k$ and other  videos in the mini-batch.
%Note that, the loss function used in Eq.~(\ref{loss}) takes all the negative triplets beyond the margin $\alpha$ into consideration, which is different from the hard negative mining used in VSE++~\cite{faghri2018vse++}.  Our experiments show that, by replacing the loss function in Eq.~(\ref{loss})  with the hard negative mining loss in VSE++~\cite{faghri2018vse++}, the performance drops. This might be due to that the hard negative mining only counts from the hardest triplet, which is prone to modal collapse.

 \subsection{Deployment}
\begin{figure}[t]
    \centering
    \includegraphics[scale=0.45]{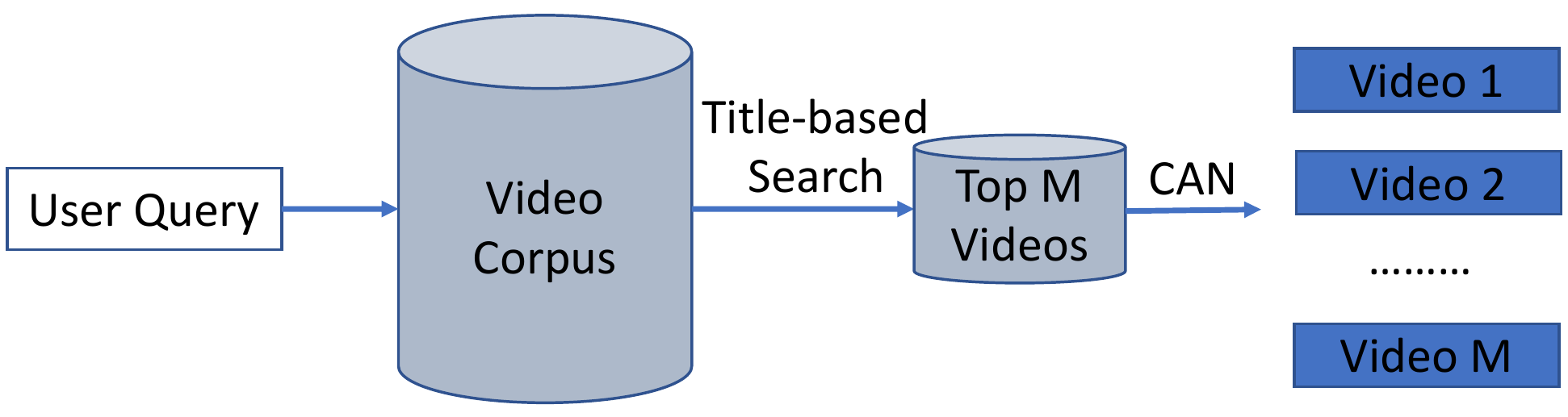}
    \caption{The deployment of CAN in the first launching. Given a query, it first uses the title-based search to recall $M$ candidate videos. After that, it uses CAN to rerank $M$ candidate videos to get the top-K videos. Since the visual information is only considered in the re-ranking phase, some relevant videos might be filtered out in the title-based search phase due to low title quality.}
    \label{rerank}
\end{figure}

Since the computational cost of the CAN is expensive, it is not practical to use it to calculate the similarity between the text query  and all videos in the database considering the number of videos is large.
 Therefore, we only deploy the CAN in the re-ranking stage.  As shown in Figure~\ref{rerank}, given a text query, we first conduct the title-based search to retrieval top M video candidates.  Since we use global feature as the title representation, it readily supports graph indexing  and thus makes the title-based search very fast.   After that,  CAN is used to re-rank the candidate $M$ videos based on their similarity scores.

  \section{Tree-based Combo-Attention Network}
 As shown in Section~\ref{cansec}, considering the efficiency issue, CAN was only deployed in the re-ranking phase for high efficiency in the first launching. But this might filter out some relevant videos before re-ranking and limits the effectiveness of CAN.  A natural question is raise: can CAN be deployed in the early stage more efficiently? One straightforward solution is to adopt the existing methods for speeding up the neural network to make CAN faster. Nevertheless, in the industrial application,  the number  of reference in the dataset are  normally in the billion scale, making the retrieval even based on a faster CAN still slow.

 In the retrieval field, a commonly used strategy to speed up the retrieval is approximated nearest neighbor (ANN) search.  By adopting some indexing methods, ANN avoids comparing the query with all reference items in the database, and thus the efficiency is significantly boosted. Nevertheless,  the existing architecture of CAN makes it incompatible with indexing-based ANN. On one hand,   CAN needs to compute the cross-modal attentions which relies on interactions between the text feature and video feature, making indexing-based ANN search not feasible. On the other hand, CAN relies on discrete local word features and bounding box features, making ANN even harder.  
%In this section, we introduce the tree-based Combo-Attention Network, which simultaneously  solves  above two  obstacles. Basically, it consists of two modules,  dual-path CAN and  tree structure. Below we introduce these two parts in details:
 In this section, we introduce the tree-based Combo-Attention Network, which simultaneously  solves  above two  obstacles. Basically, the improvement of the proposed TCAN over original CAN is three-fold: 

\begin{itemize}
\item We simultaneously consider the video's visual and title information. 
Thus, we no longer need the two-stage re-ranking process used in previous deployment of  CAN. Instead, we can directly obtain the similarity between the query and a video in a single stage.
\item We adopt a dual-path structure. On one path, it support the video/query global embedding for the further tree-based video feature indexing. On the other path, it adopts the spirit of original CAN which exploits the cross-modal attention for an effective cross-modal retrieval.
\item We build a binary search tree. The tree is constructed by the videos' embedding.  In the search phase, the query is only compared with the nodes along visiting trace, which significantly boosts the efficiency.
\item We exploit the knowledge distilling to build a lighter network for faster inference on each tree node.
%the similarity between the query and the tree node is determined by the 
\end{itemize}
 Below we introduce these four parts in details.
 %a faster CAN is still slow to conduct CAN  

\subsection{One-stage CAN} 

As shown in Figure~\ref{rerank}, the first launching of CAN adopts a two-stage method.  In the first stage, it only considers the video title and the video's visual information is considered in the second stage. In contrast, the one-stage CAN simultaneously takes the video's title and visual content into consideration.  The main difference between the original CAN and the current one-stage counterpart lies in the input.  We define the query's word features as $\mathbf{W}_q = \{\mathbf{w}_{q,1},\cdots,\mathbf{w}_{q,L}\}$, define the video's bounding box features as $\mathbf{B}_v = \{\mathbf{b}_{v,1},\cdots,\mathbf{b}_{v,M}\}$ and define the video title's word features as $\mathbf{W}_v = \{\mathbf{w}_{v,1},\cdots,\mathbf{w}_{v,M}\}$. The input of the original CAN is $[\mathbf{W}_q,\mathbf{B}_v]$. In contrast, the input of the one-stage CAN is $\mathbf{O}_{0} = [\mathbf{W}_q,\mathbf{B}_v,\mathbf{W}_v]$. The one-stage CAN also consists of a series of self-attention layers 
$\{\mathrm{SA}_{t}(\cdot)\}_{t=1}^T$. We denote the input of $i$-th SA layer by $\mathbf{O}_{i-1}$ and denote the output of the $i$-th SA layer by $\mathbf{O}_i$. In Algorithm~\ref{algself}, we give the detailed process to generate $\mathbf{O}_i$ given $\mathbf{O}_{i-1}$.

\begin{algorithm}
\caption{$i$-th self-attention layer in one-stage CAN.}
\begin{algorithmic}
  \Input:~~The input feature $\mathbf{O}_{i-1}$. 
  %\Desc{T}{matrix of measurements}
  \EndInput
  \Output:~~The  output feature  $\mathbf{O}_i$. 
  \EndOutput
\end{algorithmic}
\begin{algorithmic}[1]
  \State  $\mathbf{Q}_i= \mathbf{W}_{f_i}\mathbf{\mathbf{M}_i}$, $ \mathbf{K}_i =  \mathbf{W}_{g_i}\mathbf{\mathbf{M}_i}$,  $\mathbf{V}_i =  \mathbf{W}_{h_i} \mathbf{\mathbf{M}_i}$
  \For {$j \in [1,L+M+N]$}
      \State $\mathbf{q}_j \gets \mathbf{Q}_i[:,j]$
     \State $\mathbf{a}_j \gets \mathrm{softmax}( \mathbf{K}_i^{\top} \mathbf{q}_j)$
     \State $\mathbf{f}_j \gets \mathbf{V}_i \mathbf{a}_j$
     \State $\mathbf{o}_j = \mathrm{layernorm}(\mathbf{O}_{i-1}[:,j]+\mathbf{f}_j)$
  \EndFor
 \State $\mathbf{O}_i = [\mathbf{o}_1,\cdots, \mathbf{o}_{L+M+N}]$
   \State \textbf{return} $\mathbf{O}_i$. 
\end{algorithmic}
\label{algself}
\end{algorithm}
Using the output of the last self-attention layer $\mathbf{O}_N$, we compute the similarity between the query and the video. The first $L$ elements in $\mathbf{O}_N$ correspond to the query's local features, and the rest elements correspond to the video title's word features or bounding box features. We define $\mathbf{Q} = \mathbf{O}_N[:,0:L]$ as the query's local features, and define  $\mathbf{V} = \mathbf{O}_N[:,L:L+M+N]$ as the video's local feature. Then the similarity between the query and video is calculated by cross-matching between $\mathbf{V}$ and $\mathbf{Q}$.  In Algorithm~\ref{algsim}, we give the detailed process to generate the cross-matching similarity sim$_{\mathrm{cross}}$ given the  query's local features $\mathbf{Q}$ and the video's local features $\mathbf{V}$.
In the testing phase, the  sim$_{\mathrm{cross}}$  determines the traverse trace along the binary search tree which we will introduce later. In the training phase, sim$_{\mathrm{cross}}$ is used for constructing the triplet loss $\mathcal{L}_{cross}$.

\begin{algorithm}
\caption{Cross Matching between the query's local features $\mathbf{Q}$ and the video's local features $\mathbf{V}$.}
\begin{algorithmic}
  \Input:~~The  query's local features $\mathbf{Q}$ and the video's local features $\mathbf{V}$.
  %\Desc{T}{matrix of measurements}
  \EndInput
  \Output:~~The  cross-matching similarity sim$_{\mathrm{cross}}$. 
  \EndOutput
\end{algorithmic}
\begin{algorithmic}[1]
 
  \For {$j \in [1,L]$}
      \State $\mathbf{q}_j \gets \mathbf{Q}[:,j]$
     \State $\mathbf{a}_j \gets \mathrm{softmax}( \mathbf{V}^{\top} \mathbf{q}_j)$
     \State $\hat{\mathbf{q}}_j \gets \mathbf{V}_i \mathbf{a}_j$
  \EndFor
 \State sim$_{\mathrm{cross}} = \sum_{j=1}^L \frac{\mathbf{q}_j\hat{\mathbf{q}}_j}{\| \mathbf{q}_j \|_2 \| \hat{\mathbf{q}}_j\|_2}$
   \State \textbf{return} sim$_{\mathrm{cross}}$.
\end{algorithmic}
\label{algsim}
\end{algorithm}
%We define the 

%To be specific, in the original CAN, the input

%The only difference is the input on the video side. In original CAN, the input of the video side 
%We upgrades it to 
 
  \begin{figure}[b!]
    \centering
    \includegraphics[scale=0.45]{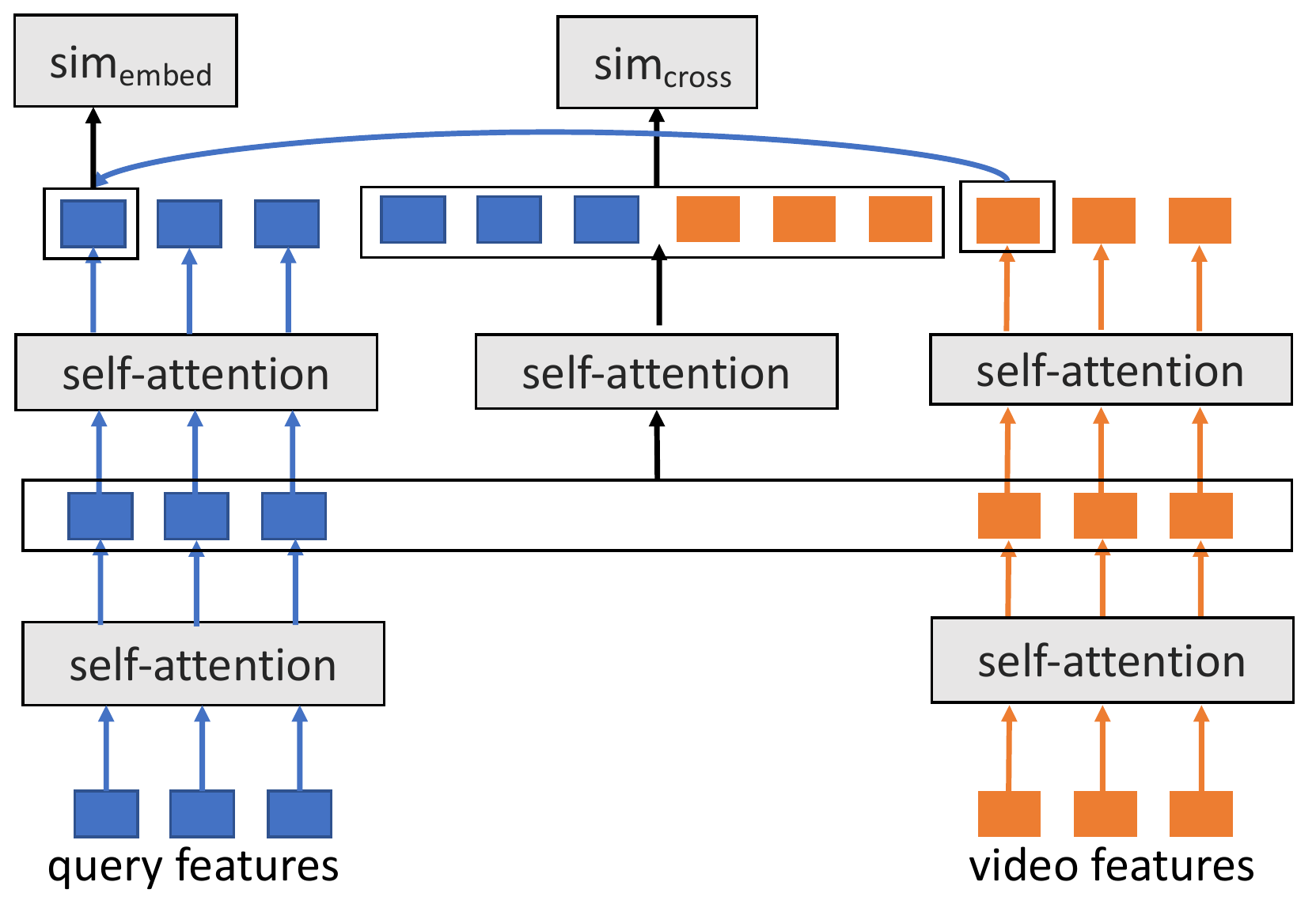}
    \caption{The architecture of dual CAN. It consists of two paths. In the embedding-path, it trains two self-attention modules for the video and query individually, and obtains the video and the query's embedding feature.  The embedding-path  is trained by the loss constructed by the embedding similarity  sim$_{\mathrm{embed}}$ calculated by the cosine similarity between two embedding features. The obtained videos' embedding features are further used to construct the binary search tree. In the cross-path, it follows the spirit of the original CAN. It concatenate the  video's local features and the query's local features, and feed them into a self-attention module to obtain the combo-attention. It is trained by the loss constructed by the cross similarity  sim$_{\mathrm{cross}}$ calculated by the soft-attention matching used in the original CAN. In the retrieval phase, sim$_{\mathrm{cross}}$  measures the relevance between the query and nodes of the binary search tree. These two paths share the same early self-attention modules, and are trained jointly. } 
    \label{dual}
\end{figure}

\subsection{Dual-path CAN} 
Dual-path CAN is extended from CAN. But we make the original CAN a dual-path structure. As visualized in Figure \ref{dual}, for the first path, it adopts the spirit of the original CAN, the query's local features and the video's local features are concatenated and fed into a self-attention module which supplies the combo-attention for an effective cross-modal retrieval.  Note that, the video's local features used in dual CAN not only includes the video's bounding box features, but also contains the video title's word features.  Meanwhile, it also adopts the same soft-attention similarity measurement as CAN based on local features  and generate the cross similarity sim$_{\mathrm{cross}}$. In the second path, two self-attention modules are trained in parallel. One self-attention module takes input the query's local features and generate the query's embedding feature $\mathbf{q}$, the first token's hidden feature of the last layer.   The other self-attention module takes input the video's local features and generates the video's global embedding feature $\mathbf{v}$. Then the similarity between the video and the query is obtained by 
\begin{equation}
\mathrm{sim}_{\mathrm{embed}} = \langle\mathbf{q},\mathbf{v} \rangle / \|\mathbf{q}\| \|\mathbf{v}\|. 
\end{equation} 
When training the dual-path CAN, we compute a triplet loss $\mathcal{L}_{\mathrm{cross}}$ based on $\mathrm{sim}_{\mathrm{cross}}$ and the other triplet loss $\mathcal{L}_{\mathrm{embed}}$ based on $\mathrm{sim}_{\mathrm{embed}}$. The  loss function is obtained by a weighted summation of these two losses:
\begin{equation}
\mathcal{L}_{dual} = \mathcal{L}_{\mathrm{cross}}  + \lambda \mathcal{L}_{\mathrm{embed}},
\end{equation}
where $\lambda$ is a positive constant which we set as $0.5$ by default. 
In the search phase, we only use the cross similarity for the tree traverse.  On the other hand, we use the video embedding trained from the $\mathcal{L}_{\mathrm{embed}}$ to construct the tree.

\subsection{Tree Structure}
The binary search  tree is constructed based on videos' embedding   feature $\{\mathbf{v}_1,\cdots,\mathbf{v}_N\}$.  
To be specific, we use hierarchical k-mediods to build the binary search tree based on the clustering results of videos' embedding   feature $\{\mathbf{v}_1,\cdots,\mathbf{v}_N\}$. 
Note that, there are two ways to conduct the hierarchical k-mediods, the top-down way and the bottom-up way. We use the top-down way  considering the efficiency.  
Meanwhile, we use the industry label of each sample as the initial clustering label for k-mediods.  In the tree, the nodes in each layer corresponds to mediods in that level. Meanwhile, we use the node videos' local features  to compute the cross similarity between the query and the node video.

The tree structure and the dual-path CAN are trained in an alternating manner by updating one and fixing the other. To be specific, it conducts two steps alternately. In the first step,  for each query,  we sample several negative nodes from each layer to train the dual-path CAN model. In the second step, it updates the tree structure based on the videos' embeding features obtained from the first step.  Note that, the negative sampling  strategy  is quite important for the performance of the proposed tree-based CAN. In Experiment Section, we will compare different negative sampling in details.

\begin{algorithm}[h]
\caption{The  tree-based top K retrieval.}
\begin{algorithmic}
  \Input:~~~ The query's embedding feature $\mathbf{q}$, the query's word feature set $\mathcal{W} = \{\mathbf{w}_1,\cdots,\mathbf{w}_M\}$,  the number of retrieved videos $K$, the trained tree with depth $L$ and the learned dual-path CAN. 
  %\Desc{T}{matrix of measurements}
  \EndInput
  \Output:~~~ The top $K$ retrieved videos. 
  \EndOutput
\end{algorithmic}
\begin{algorithmic}[1]
  \State Candidate set $\mathcal{C} = \{\mathrm{root}~\mathrm{video}~\mathrm{node} \}$. 
  \For {$i \in [1,L]$}
      \For {$ \mathcal{V} \in \mathcal{C}$}
         \State  calculate sim$_{\mathrm{cross}}$($\mathcal{V}, \mathcal{W} $)
     \EndFor
     \State   sort  $ \mathcal{V} \in \mathcal{C}$ in descending order of sim$_{\mathrm{cross}}$($\mathcal{V}, \mathcal{W} $)
     \State   get top $K$ video nodes $\mathcal{P}$.
     \State  $\mathcal{C} =\{ \mathrm{children~nodes~of} \mathcal{V} | \mathcal{V} \in \mathcal{P} \} $ 
  \EndFor

   \State \textbf{return} top $K$ items in $\mathcal{C}$ based on the sim$_{\mathrm{cross}}$($\mathcal{V}, \mathcal{W} $).
\end{algorithmic}
\label{alg1}
\end{algorithm}

Algorithm~\ref{alg1} summarizes the tree-based top K retrieval process. Benefited from the tree structure, the computation complexity of the tree-based top-K retrieval is only $\mathcal{O}(N)$ where $N$ is the number of videos in the database. To be more specific, for a dataset consisting of $10$ million videos,  it builds a $23$-layer binary search tree. For each layer, we needs compute sim$_{\mathrm{cross}}$ for two nodes.  Thus, in total, we only need visit $1+22\times 2 = 45$ nodes, \emph{i.e.}, computing $45$ times sim$_{\mathrm{cross}}$ to find the top 1 video.

  \begin{figure*}[t]
    \centering
    \includegraphics[scale=0.45]{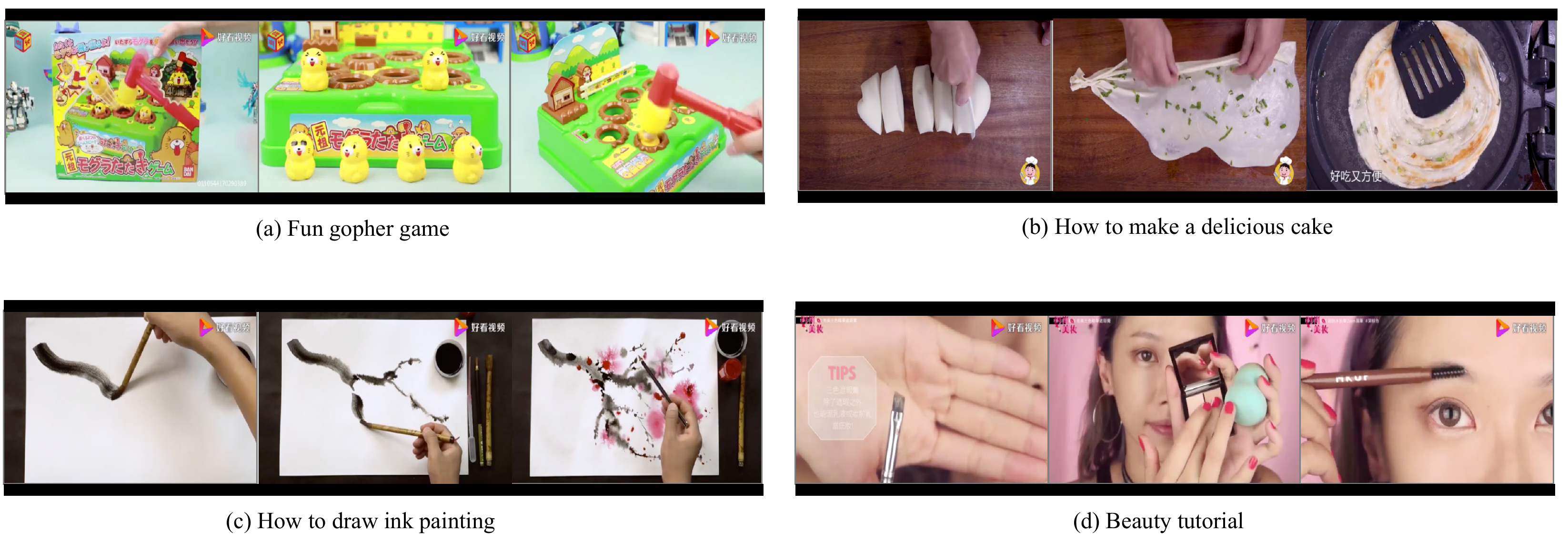}

\vspace{-0.1in}

    \caption{The visualization of query-video pairs. In our Daily1.2B, texts are in Chinese. For the convenience of illustration, we translate the Chinese to English.}
    \label{queries}\vspace{-0.1in}
\end{figure*}

\subsection{Knowledge Distilling}
Considering the heavy computation cost in self-attention layers for evaluating the tree node. We further improves the efficiency through knowledge distilling. We design a student network consists of two self-attention layers with hidden size as $256$ to distill the knowledge from $4$ self-attention layers with hidden size as $768$. The distilling loss is designed in the following way:
\begin{equation}
\mathcal{L}_{distill} = \mathcal{L}_{\mathrm{cross}_{\mathrm{std}}} +  \gamma \mathrm{MSE}(\mathrm{sim}_{\mathrm{cross}} - \mathrm{sim}_{\mathrm{cross}_{\mathrm{std}}} ),
\end{equation}
where $\mathrm{MSE}$ denotes the mean square error,  $\mathcal{L}_{\mathrm{cross}_{\mathrm{std}}}$ corresponds to the triplet loss computed based on the cross path of the student network, and $\mathrm{cross}_{\mathrm{std}}$ corresponds to the cross similarity obtained from the student network. $\gamma$ is a constant positive, which we set as $0.3$ by default.
The final loss is defined as a weighted summation  of $\mathcal{L}_{dual}$ and $\mathcal{L}_{distill}$:
\begin{equation}
\mathcal{L} = \mathcal{L}_{dual} + \beta \mathcal{L}_{distill},
\end{equation}
where $\beta$ is a positive constant we set as $0.5$.

 % It 

\section{Off-line Experiments}

\subsection{Datasets and Implementation Details}
The experiments are conducted on our collected short video dataset, Daily1.2B. It consists of $1.2$ billion pairs of short videos and query sentences, which are mainly about daily lives collected from Baidu's Haokan APP. Since the relevance between a query and the short videos is relatively subjective to users, we collect ground-truth pairs by  selecting  the query-video pairs with high click rates, representing good ones for a large number of users. In Figure~\ref{queries}, we visualize some pairs of short videos and query sentences.

For each video, we sample $16$ key frames from each video. For each key frame, $32$ bounding boxes are generated from Faster R-CNN~\cite{ren2015faster} built on ResNet-101~\cite{he2016deep} pre-trained on Visual Genomes~\cite{krishna2017visual}. For each detected region of the interest (ROI), \emph{i.e.}, the bounding box, its feature is obtained by sum-pooling the convolutional features within the bounding box. The feature dimension is $2048$. We use k-means to group $16\times32 = 512$ detected bounding boxes into $32$ clusters and use the cluster centroids as the video's initial representation. We set the number of head in combo-attention network as $8$.    The tree contains 8 million nodes in total, among which, 4 million are leaf nodes. % We use Adam as the optimizer.
All models are trained and deployed based on the PaddlePaddle deep learning framework developed by Baidu.

%In the  global-feature method,  we obtain the global feature of the video through sum-pooling the all  hidden vectors  of the BiLSTM. 

%Since the 
%Then a new similarity matrix $\hat{\mathbf{S}$
%\subsection{Sentence Representation}

%Recall from Figure~\ref{basicmodule}(b) that, the combo-attention block consists of a self-attention module and a cross-modal attention module. As shown in Table~\ref{casetting}, by removing the cross-modal attention module,  s2v recall@1  decreases from $68.8$  to $61.4$ and v2s recall@1 decreases from $68.8$ to $61.3$, validating the effectiveness of the cross-modal attention module.

%Meanwhile, removing the self-attention module, the  sentence-to-video recall@1  decreases  to $26.5$ and  video-to-sentence recall@1 decreases  to $23.2$. 

%\textbf{Influence of spatial locations and temporal locations.}  Recall from Eq.~(\ref{sptemp}) that, a bounding box feature is obtained by summing up the visual feature and the spatial and temporal locations.  We investigate the influence of the spatial and temporal locations. As shown in Table~\ref{st}, without spatial and temporal 
%locations, it only achieves $28.2$ s2v recall@1. By adding the spatial locations, s2v recall@1 increases to $29.3$. However, when adding both spatial and temporal locations, s2v recall@1  drops to $28.7$. By default, we only add the spatial location for each bounding box.

\subsection{One-stage CAN.}

\textbf{Comparisons with two-stage CAN.} As we mentioned previously, in the first launching, we conduct a two-stage retrieval process. In this first stage, it exploits title-based search to conduct the coarse-level search and then use CAN for reranking.  We compare the proposed one-stage CAN with the two-stage baseline.  We vary the number of candidate item pool $M$, among $\{32,64,128,256,512\}$. Two-stage baseline first conduct the title-based search to retrieve the top $20$ items and then conduct re-ranking based on the original CAN.  In contrast, our one-stage CAN directly calculate the similarity between the query and $M$ candidate items using the video's title and visual information. We compare the mAP@3 achieved by ours and that based on two-stage search. As shown in Table~\ref{map3}, our one-stage CAN consistently outperforms two-stage CAN. %For instance, when $M=256$,  the retrieval  based on two-stage CAN only achieves a $0.948$ mAP@3, whereas our one-stage CAN achieves a $0.971$ mAP@3.

\begin{table}[htp]
\centering
\begin{tabular}{c|c|c|c|c|c}
\hline
$M$     & $32$ & $64$ & $128$ & $256$   & $512$ \\ \hline
two-stage & $0.984$   &   $0.975$  & $0.962$ &$0.948$& $0.928$  \\ \hline
one-stage & $0.993 $   &   $0.989$   & $0.980$  &$0.971$& $0.950$ \\ \hline
\end{tabular}
\vspace{5mm}
\caption{mAP@3 comparisons between TCAN with title-based  search.}
%\vspace{5mm}
\label{map3}
\end{table}

\textbf{Ablation Study.} The one-stage CAN simultaneously considers the video's title and visual content.  We study the influence of each part on the retrieval performance.
As shown in Table~\ref{AS}, using only video title's word features, it only achieves a $0.972$ mAP@3. In contrast, using only video's visual features, the bounding box features, it only achieves a $0.912$ mAP@3. Both of them are lower than mAP@3 achieved by taking both title and visual features into consideration.  Meanwhile, despite that there are a large performance gap between the title local features and bounding box visual features, fusing them still achieves a better performance. It demonstrates the effectiveness of self-attention layers in feature fusion.  

\begin{table}[htp]
\centering
\begin{tabular}{c|c|c|c|c|c}
\hline
$M$     & $32$ & $64$ & $128$ & $256$   & $512$ \\ \hline
title & $0.972$   &   $0.960$  & $0.952$ &$0.938$& $0.916$  \\ \hline
visual & $0.912 $   &   $0.903$   & $0.892$  &$0.881$& $0.872$ \\ \hline
title $\&$ visual   & $0.993 $   &   $0.989$   & $0.980$  &$0.971$& $0.950$ \\ \hline
\end{tabular}
\vspace{5mm}
\caption{Ablation study on the influence of videos' titles and visual content on the retrieval performance. The performance is evaluated by mAP@3.}
%\vspace{5mm}
\label{AS}
\end{table}

\subsection{Knowledge Distilling}

We conduct the ablation study on knowledge distilling. As mentioned, the original CAN uses $4$ self-attention layers with hidden size as $768$. In contrast, the student network only uses two self-attention layers with hidden size as $256$. As shown in Table~\ref{distill}, the performance achieved by the student network is comparable with that of the original CAN, demonstrating the effectiveness of the knowledge distilling. For instance, the original CAN achieves a $0.698$ mAP@1, whereas the student network achieves a $0.694$ mAP@1.

\begin{table}[h]
\centering
\begin{tabular}{c|c|c|c}
\hline
  & \multicolumn{3}{c}{mAP@}   \\\cline{2-4} 
             \multirow{2}{*}                     & 1    & 3   & 5      \\ \hline
    Original     &   $0.698$     &   $0.731$    &   $0.791$         \\ \hline
    Student     &   $0.694$     &   $0.725$    &   $0.780$       \\ \hline 
\end{tabular}
\vspace{5mm}
\caption{The ablation study on knowledge distilling. }
%\vspace{5mm}
\label{distill}
\end{table}

\subsection{Comparison with embedding-based  binary search tree.}  An alternative solution is to use the cosine similarity between query's embedding  and the video's embedding obtained from the embedding-path to replace the similarity calculated from the CAN when traversing the binary search tree.  We compare the TCAN with embedding-based  baseline. 
As shown in Table~\ref{emb}, the proposed TCAN consistently outperforms the embedding-based  binary search tree. For instance, our TCAN achieves a $0.694$ mAP@1, whereas mAP@1 of the embedding-based  binary search tree is only $0.627$.

\begin{table}[h]

\vspace{0.2in}

\centering
\begin{tabular}{c|c|c|c}
\hline
  & \multicolumn{3}{c}{mAP@} \\\cline{2-4} 
             \multirow{2}{*}                     & 1    & 3   & 5      \\ \hline
     Embed     &   $0.627$     &   $0.661$    &   $0.746$         \\ \hline
      TCAN      &   $0.694$     &   $0.725$    &   $0.780$       \\ \hline 
\end{tabular}
\vspace{5mm}
\caption{The comparisons among  negative sampling strategies. }
\label{emb}\vspace{-0.2in}
\end{table}

\begin{figure*}[t]
\centering
    \subfigure[The practice of tomato fish.]{\includegraphics[width=0.39\textwidth,height=0.32\textwidth]{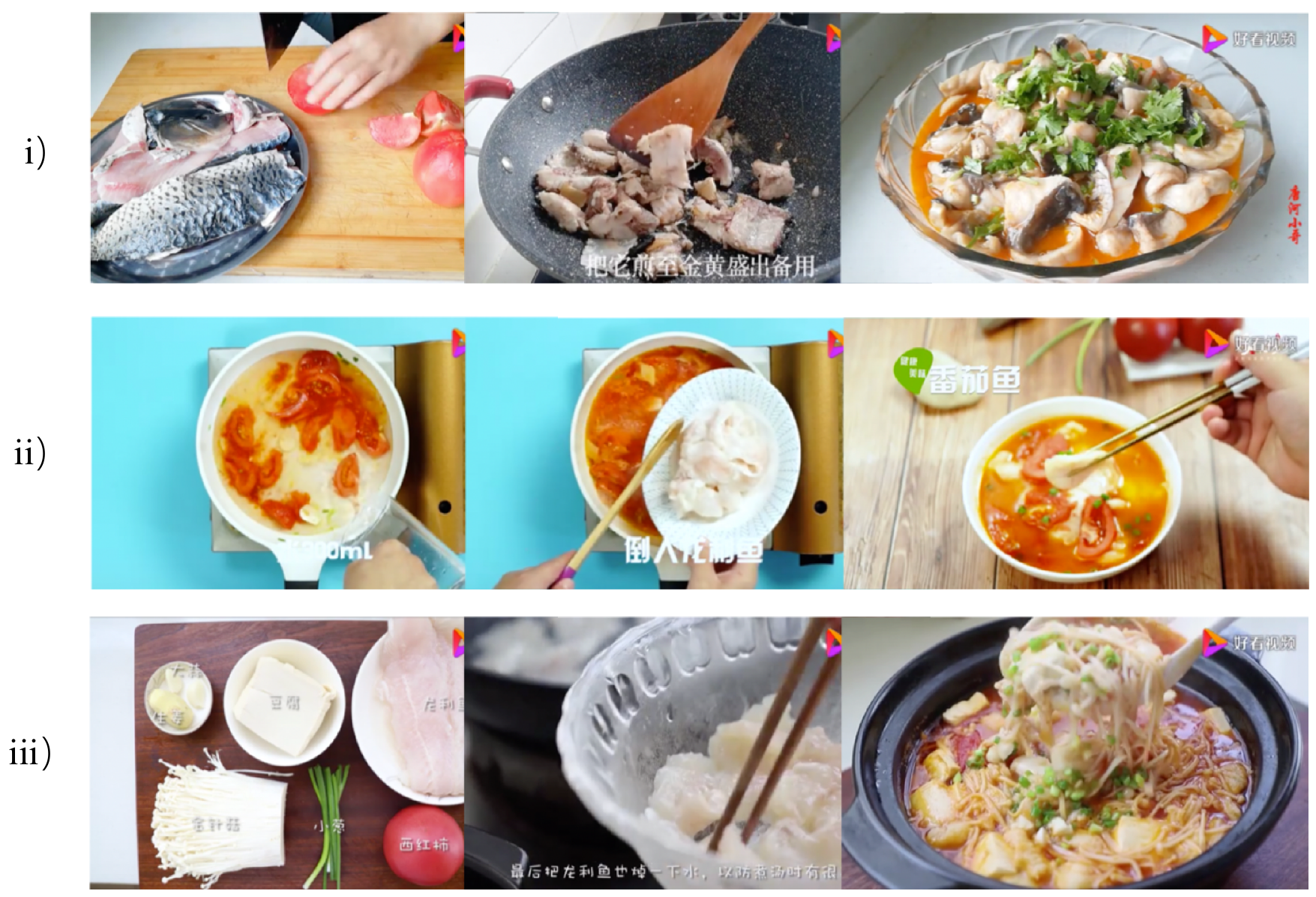}}  
    \hspace{1mm}
    \subfigure[How to draw an elephant stick figure.]{\includegraphics[width=0.39\textwidth,height=0.32\textwidth]{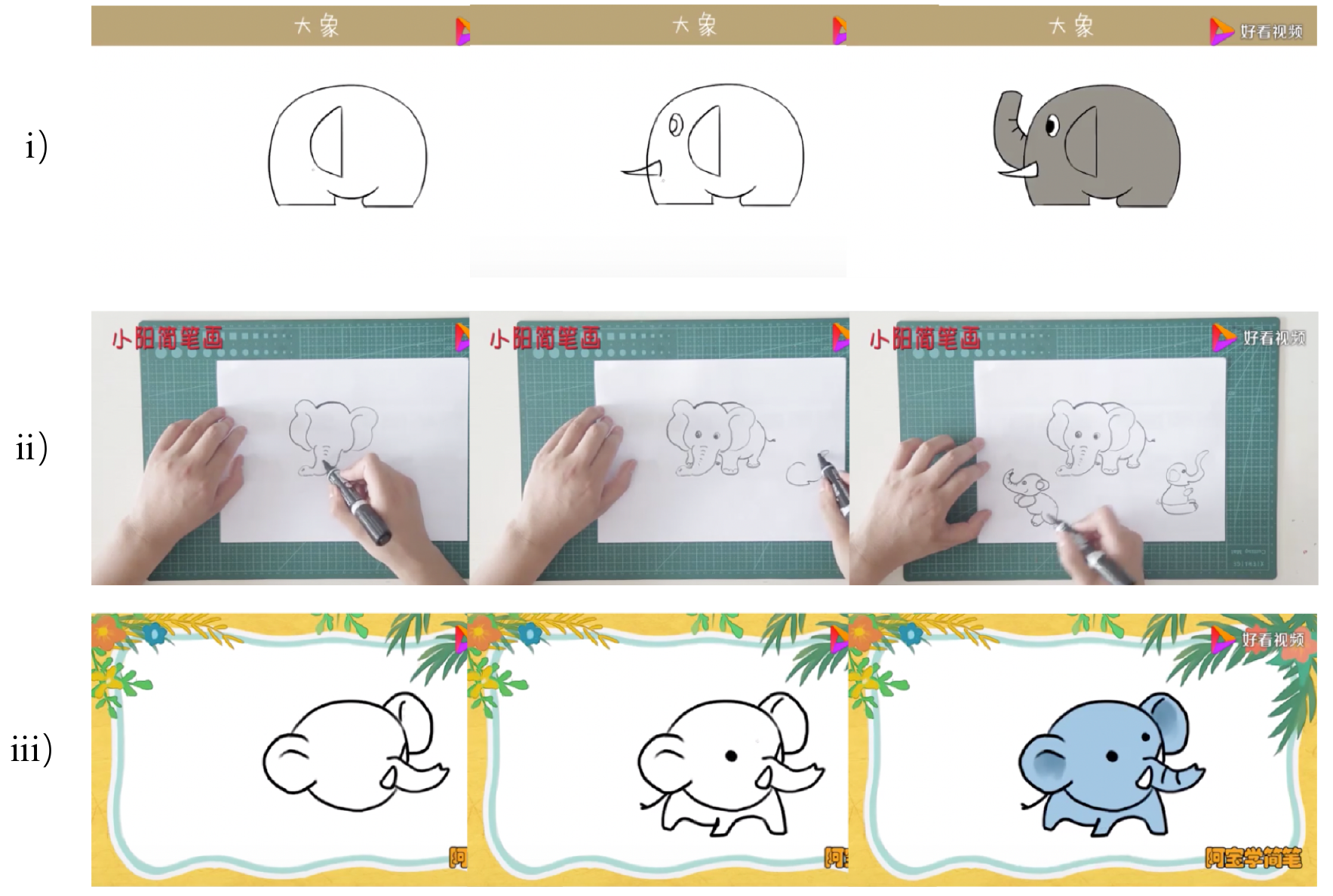}}  
      \subfigure[Introduction video of Mercedes-Maybach.]{\includegraphics[width=0.392\textwidth,height=0.32\textwidth]{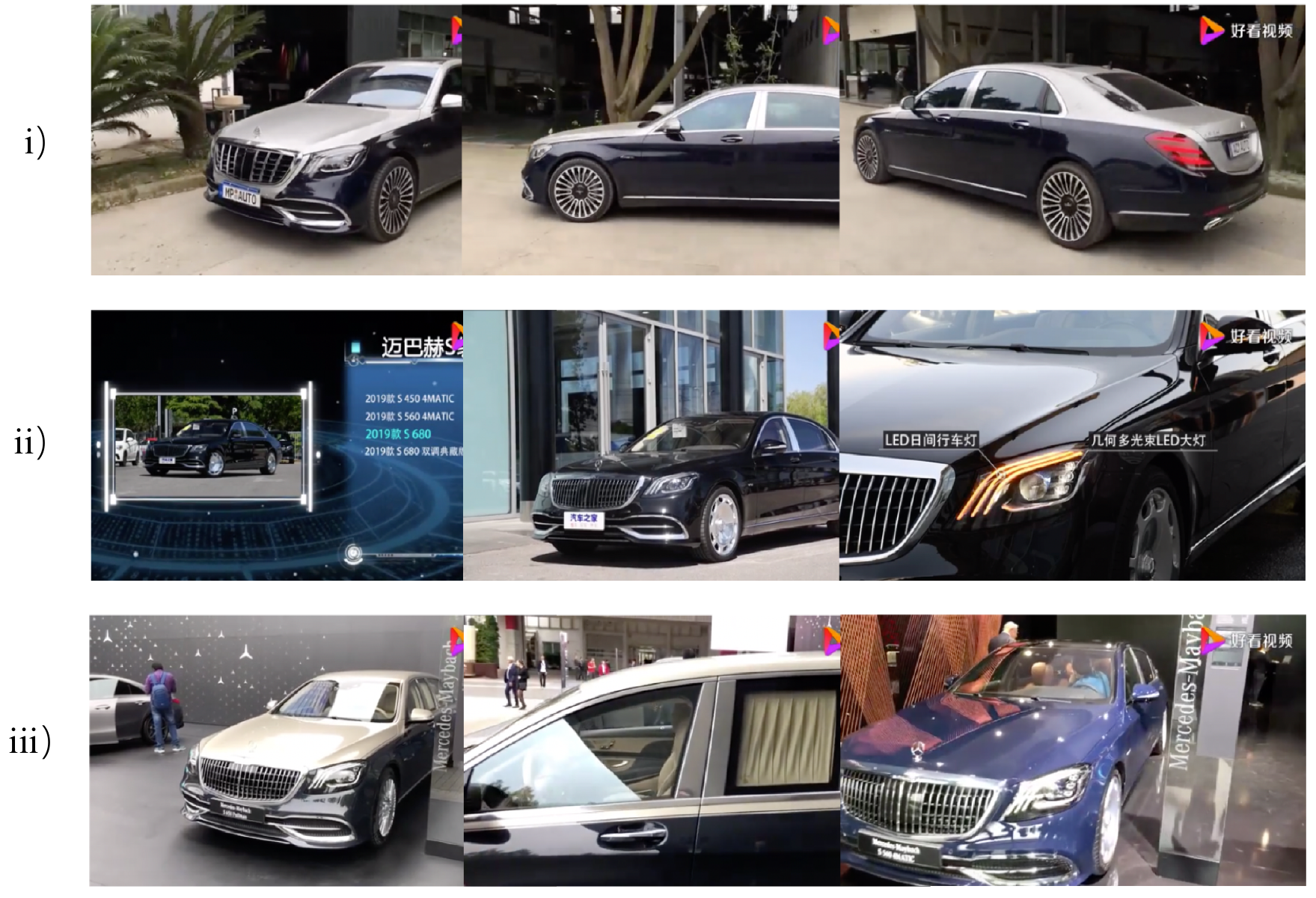}}  
        \subfigure[Tai Chi Teaching Video.]{\includegraphics[width=0.388\textwidth,height=0.32\textwidth]{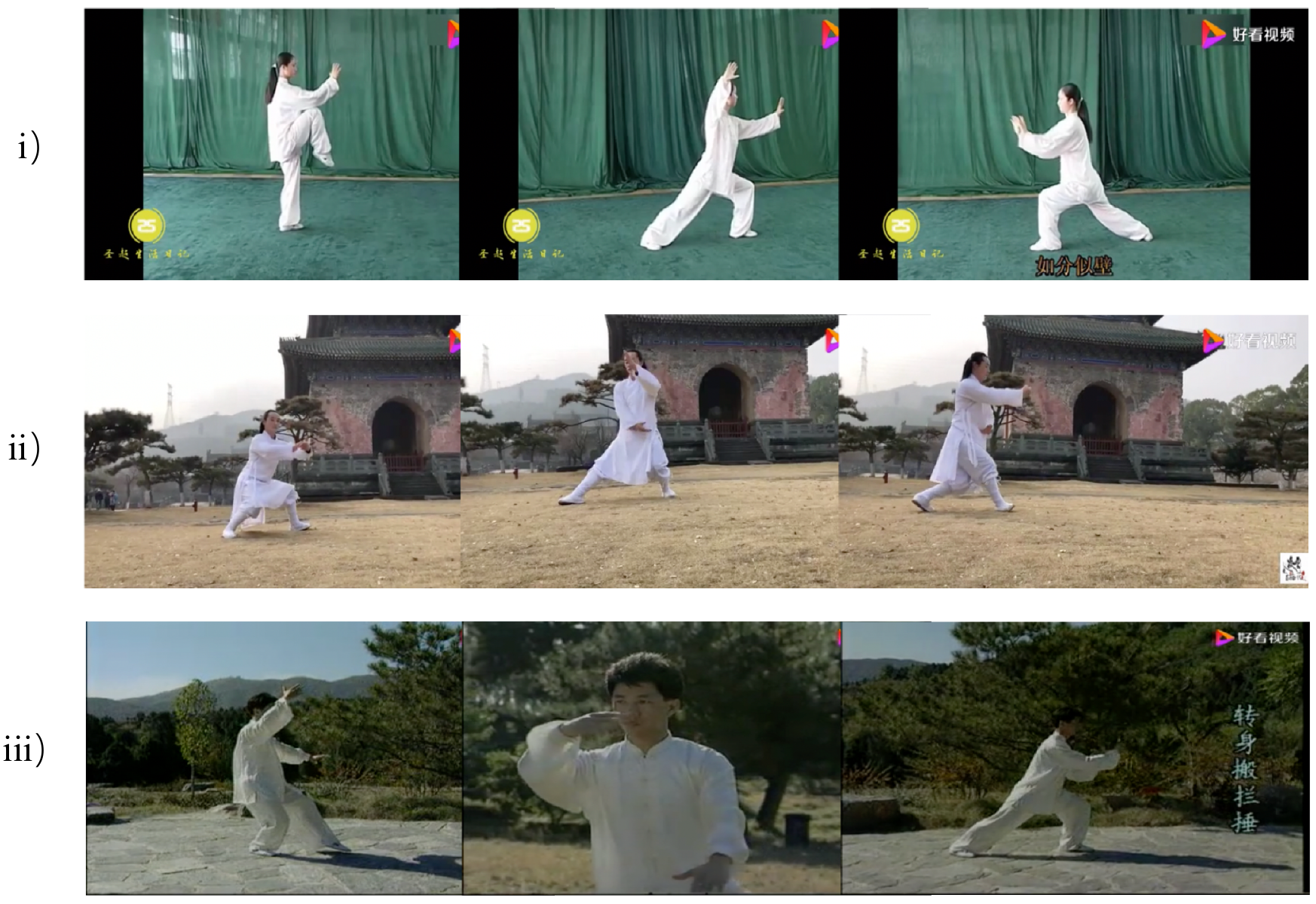}}  
    \caption{Visualization of top-3 retrieved videos.  Note that, in our Daily1.2B dataset, the texts are in Chinese. }
    \label{resultvisual}\vspace{-0.1in}
\end{figure*}

\begin{table}[b!]
\centering
\begin{tabular}{c|c|c|c|c|c}
\hline
\multirow{2}{*}{uniform} & \multirow{2}{*}{arithmetic} & \multicolumn{4}{c}{geometric} \\ \cline{3-6} 
                         &                              & 1.2    & 1.3   & 1.4   & 1.5   \\ \hline
                    $0.713$     &          $0.821$                    &   $0.860$     &   $0.895$    &   $0.966$    &  $0.965$     \\ \hline
\end{tabular}
\vspace{1mm}
\caption{The comparisons among  negative sampling strategies. }
%\vspace{1mm}
\label{auc}
\end{table}

\subsection{Negative sampling strategy} As we mentioned, when training the dual-path CAN model, we sample negative nodes from each layer of the tree. Since the number of nodes increases exponentially as the depth increases, a general guidance for negative sampling is to sample more negative nodes in deeper layers. We compare different negative sampling strategies including uniform sampling, arithmetic sampling, and geometric sampling.  In uniform sampling, we sample the same number of negative nodes for each layer. In arithmetic sampling, we sample $L$ negative samples for the $L$-th level. In geometric sampling, we sample $\lceil L^{\alpha} \rceil $ samples  for the $L$-th level. We test the performance of  geometric sampling when $\alpha \in \{1.2, 1.3, 1.4, 1.5\}$. The evaluation metric is AUC of the precision-recall retrieval result. As shown in Table~\ref{auc}, the negative sampling strategies have significant influence on the retrieval performance.  To be specific, the AUC achieved by uniform sampling is only $0.713$, and that achieved by arithmetic sampling is only $0.821$. In contrast, geometric sampling with $\alpha=1.4$ achieves s $0.966$ AUC. By default we adopt geometric sampling with $\alpha=1.4$.

Figure~\ref{resultvisual} visualizes the retrieval result. For each query, we show three key frames of top3 retrieved videos. As shown in the figure, the retrieval quality is quite high.

\section{Online Experiments}

We evaluate the proposed TCAN in Baidu dynamic video advertising platform. Two online metrics are used to measure the performance:  click-through rate (CTR)  and conversion rate (CVR)  defined as follows:
\begin{equation}
 \mathrm{CTR} = \frac{ \# ~  \mathrm{of}~\mathrm{clicks}}{ \# ~  \mathrm{of}~\mathrm{impressions}}, \mathrm{CVR} = \frac{\mathrm{revenue}}{ \# ~  \mathrm{of}~\mathrm{clicks}}.
\end{equation}
We compare the CTR and CVR of  Baidu dynamic video advertising platform before and after launching the proposed TCAN. Note that, before launching TCAN, the video search is based on title-based retrieval followed by CAN reranking.

\newpage

\begin{table}[t]
\centering
\begin{tabular}{c|c|c}
\hline
metric      & CTR & CVR \\ \hline
improvement & $2.29\%$   &   $2.63\%$    \\ \hline
\end{tabular}
\vspace{5mm}
\caption{Online results from May. 17th to May. 20th, 2020 in Baidu dynamic video advertising platform.}
%\vspace{5mm}
\label{online}
\end{table}

 As shown in Table~\ref{online}, after launching TCAN, CTR achieves a $2.29\%$ increase and CVR achieves a $2.63\%$ increase.

%\begin{table}[htp]
%\begin{tabular}{c|c|c|c|c}
%\hline
%       & \multicolumn{4}{c}{Recall@} \\ \hline
%method & $1$    & $5$    & $10$    & avg    \\ \hline
%VSE++~\cite{faghri2018vse++}   &  -    &   -  &  -     &     -   \\ \hline
%SCAN~\cite{lee2018stacked}   &  $-$    &   $-$   &  $-$     &     $-$   \\ \hline
%CAN (Ours)   & $-$    & $-$      & $-$  &$-$         \\ \hline
%\end{tabular}
%\caption{Comparisons with SCAN~\cite{lee2018stacked} on  Daily600K and VATEX~\cite{wang2018videos} datasets.}
%\label{cpstoa}
%\end{table}

\vspace{-0.1in}
\section{Conclusion}
In this paper, we present the tree-based combo-attention network (TCAN) recently launched in Baidu dynamic video adverting platform.  By extending the original CAN to dual-path CAN, it simultaneously supports the cross-modal attention as well as the global feature embedding.  Based on the proposed dual-path CAN, we build a binary search tree, which effectively avoid the exhaustive search and significantly boost the retrieval efficiency. The systematic  experiments conducted on offline experiments demonstrate its effectiveness for cross-modal retrieval. Meanwhile, the online experiments show the launch of TCAN considerable boosts the revenue of Baidu dynamic video adverting platform. 

% Despite the combo-attention network has significantly improved the performance 

%\begin{acks}
%We are deeply grateful to the contributions of many colleagues from Baidu. A few names are Shujing Wang, Xiaodong Chen, Lin Liu, Zhiqiang Xu; but there are certainly many more who have contributed to this large project.
%\end{acks}

\bibliographystyle{plain}
\bibliography{refs_scholar}

\end{document}